\documentclass[10pt,twocolumn,letterpaper]{article}

\usepackage{iccv}
\usepackage{times}
\usepackage{epsfig}
\usepackage{graphicx}
\usepackage{amsmath}
\usepackage{amssymb}
\usepackage{authblk}
\usepackage{mathtools}
\usepackage{times,microtype}
\usepackage{multirow}
\usepackage[ruled,vlined]{algorithm2e}
\usepackage{array}
\usepackage{booktabs}
\usepackage{caption,balance}
\usepackage{xspace,colortbl}
\usepackage{listings}
\usepackage{enumitem}
\usepackage{multirow,microtype}
\usepackage{subcaption,microtype}
\usepackage{mathtools}
\usepackage{xcolor}
\usepackage{cite}
\usepackage[flushleft]{threeparttable}
\usepackage[export]{adjustbox}
\newcommand{\smoco}{\textsc{SupMoCo}\xspace}
\newcommand{\supcon}{\textsc{SupCon}\xspace}
\newcommand{\md}{\textsc{Meta-Dataset}\xspace}
\newcommand{\moco}{\textsc{MoCo}\xspace}
\newcommand{\simclr}{\textsc{SimCLR}\xspace}

\definecolor{backcolour}{rgb}{255,255,255}

\lstdefinestyle{mystyle}{
    backgroundcolor=\color{backcolour},   
    commentstyle=\color{blue},
    keywordstyle=\color{red},
    stringstyle=\color{codepurple},
    basicstyle=\ttfamily\scriptsize,
    breakatwhitespace=false,         
    breaklines=true,                 
    captionpos=b,                    
    keepspaces=true,                 
    showspaces=false,                
    showstringspaces=false,
    showtabs=false,                  
    tabsize=2
}

\definecolor{citecolor}{HTML}{0071bc}
\usepackage[pagebackref=true,breaklinks=true,letterpaper=true,colorlinks,citecolor=citecolor,bookmarks=false]{hyperref}

\iccvfinalcopy 

\begin{document}

\title{Supervised Momentum Contrastive Learning for Few-Shot Classification}

\author[1]{Orchid Majumder\thanks{correspondence to: orchid@amazon.com}}
\author[1]{Avinash Ravichandran}
\author[2]{Subhransu Maji}
\author[1]{Alessandro Achille}
\author[1]{Marzia Polito}
\author[1,3]{Stefano Soatto}
\affil[1]{Amazon Web Services}
\affil[2]{UMass Amherst}
\affil[3]{UCLA}

\maketitle

\begin{abstract}
Few-shot learning aims to transfer information from one task to enable generalization on novel tasks given a few examples. 
This information is present both in the domain and the class labels.
In this work we investigate the complementary roles of these two sources of information by combining instance-discriminative contrastive learning and supervised learning in a single framework called \textbf{Sup}ervised \textbf{Mo}mentum \textbf{Co}ntrastive learning (\smoco).
Our approach avoids a problem observed in supervised learning where information in images not relevant to the task is discarded, which hampers their generalization to novel tasks.
We show that (self-supervised) contrastive learning and supervised learning are mutually beneficial, leading to a new state-of-the-art on the ~\md~\cite{triantafillou2019meta} --- a recently introduced benchmark for few-shot learning.
Our method is based on a simple modification of \moco~\cite{he2020momentum} and scales better than prior work on combining supervised and self-supervised learning.
This allows us to easily combine data from multiple domains leading to further improvements.
\end{abstract}

\section{Introduction}

A few-shot learning system should learn a representation of the data that is invariant to common factors of variations of objects (e.g., change of pose, deformations, color) while still representing features that allow to discriminate between different classes. Factoring out all the nuisance factors reduces the effective dimensionality of the hypothesis space and allows to learn good classifiers using only a few samples. For this reason, much of the few-shot literature hinges on the intrinsic ability of deep neural networks (DNNs) to learn invariant representations when trained in a supervised manner. However, DNNs are often too eager to learn invariances. In what is known as "supervision collapse"~\cite{doersch2020crosstransformers}, a DNN can learn to encode only the features that are useful to discriminate between the training classes, and as a result is not sufficiently expressive to discriminate between new unseen classes which is what eventually matters in few-shot learning. The question is then: How can we learn a representation that is invariant to common factors while maintaining discriminativeness for unseen classes?

\begin{figure}
    \includegraphics[width=1.0\linewidth]{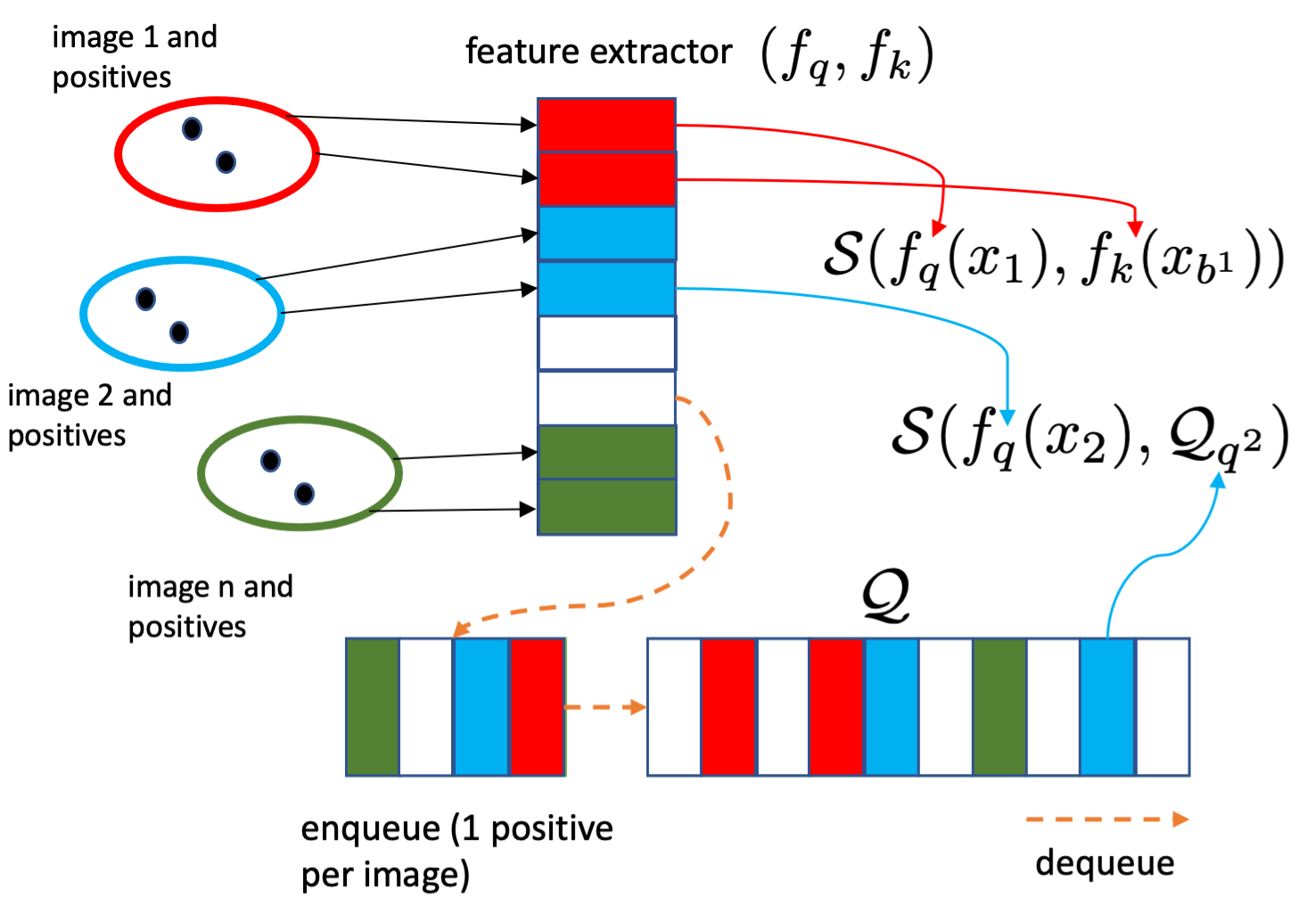}
    \caption{\small High-level illustration of \smoco (Sec.~\ref{sec: supmoco}). During training, for each image, we collect $P$ additional images (referred as positives, $P=1$ in the figure) out of which one is the augmented view of the image and the rest are random augmented samples from the same class. The original image is fed through the query-encoder ($f_q$) where the other images goes through the key-encoder ($f_k$) for feature extraction. Once feature extraction is complete, we use an instance-discriminative contrastive loss to maximize similarity between the features of the image and its positives. Apart from these $P$ positives, we also identify entries belonging to the same class from the queue ($\mathcal{Q}$) (used to store features of past samples) and maximize similarity with those as well for every image. Features corresponding to the data augmented view of each sample are inserted into the queue and the oldest entries are removed.}
    \label{fig:moco}
\end{figure}

In this paper we introduce  \textbf{Sup}ervised \textbf{Mo}mentum \textbf{Co}ntrastive learning (\smoco). \smoco (Fig.~\ref{fig:moco}) augments standard self-supervised learning to account for class labels, so that the network learns the intra-class variability through supervision while at the same time retaining distinctive features of the individual images through the self-supervised components, thus avoiding supervision collapse. On the algorithmic side, \smoco makes extensive use of the efficient queue based architecture of \moco, which avoids memory bottleneck and leads to a greater diversity of classes in the contrastive objective. We found this to be critical for good performance, and it allows \smoco to achieve a significantly better performance than other comparable algorithms (\cite{khosla2020supervised, doersch2020crosstransformers}) in the literature. On the popular \md~\cite{triantafillou2019meta} few-shot benchmark, \smoco achieves a new state-of-the-art (Tab.~\ref{Tab:md-im},~\ref{Tab:md-all}) and we observe an average $4\%$ accuracy increase  (Tab.~\ref{Tab:cs-im},~\ref{Tab:cs-all}) over the closest comparison (\supcon \cite{khosla2020supervised}) .

\smoco allows us to easily combine data from different domains during training in a multi-domain setup. 
Compared to training on a single domain (ImageNet), training on a combination of domains leads to a large improvement in performance on novel tasks where the domain difference from ImageNet is large (\eg, Quickdraw, Aircraft, and Fungi) as seen in Tab.~\ref{Tab:md-im}, \ref{Tab:md-all}.
In a partially labeled setup where we provide all labeled samples from ImageNet and only $10\%$ of labeled data from the remaining ones with the rest provided as unlabeled, \smoco only suffers an average $2\%$ performance drop (Tab.~\ref{Tab:ssl}) and beats several recently proposed few-shot learning algorithms using all supervision.

We perform an ablation study to investigate the complementary roles of supervised and self-supervised learning by analyzing the degree of generalization and supervision collapse (Fig.~\ref{fig:collapse}). We visualize the distribution of nearest neighbors obtained through representations trained using supervision and with \smoco using the empirical framework presented in~\cite{doersch2020crosstransformers}.
We find that \smoco avoids supervision collapse better than the supervised method in this experiment.

\section{Related Work}

\noindent
\textbf{Few-shot learning methods.} Meta-learning and standard supervised learning have been the two most common approaches for pre-training a representation for few-shot classification. 
Meta-learning methods can be broadly classified into metric learning based and optimization based techniques. 
Metric learning \cite{snell2017prototypical, Koch2015SiameseNN, oreshkin2018tadam, vinyals2016matching, ravichandran2019few, sung2018learning, wang2019simpleshot} methods learn a feature representation such that similar images are close in the embedding space relative to dissimilar images. 
Optimization based methods \cite{finn2017model, lee2019meta, bertinetto2018meta} learn representations that lead to generalizable models measured using pre-defined classification model, objective, or a training procedure.
On the other hand \cite{chen2019closer, dhillon2019baseline, wang2019simpleshot, chen2020new, tian2020rethinking} showed that competitive performance can be obtained using standard cross-entropy based training with a few modifications, suggesting the need to understand the conditions under which models are transferable.
This is the focus of a broader class of meta-learning approaches that aim to improve few-shot transfer through techniques for model and dataset selection, designing task representations, and modeling transferability (\eg,~\cite{achille2019task2vec,Wang_2019_CVPR,yan2020neural,tran2019transferability,zamir2018taskonomy}).

\noindent
\textbf{Few-shot learning benchmarks.} Popular few-shot benchmarks such as miniImageNet \cite{vinyals2016matching} and tieredImageNet~\cite{oreshkin2018tadam} divide the ImageNet dataset~\cite{russakovsky2015imagenet} into a disjoint train, validation, and test set of classes. The train set of classes are used for pre-training and few-shot evaluation is performed on tasks sampled from the test set varying the number of classes and labeled examples (\eg, $5$-way-$5$-shot). 
These benchmarks exhibit relatively small domain shifts.
As an alternate \cite{triantafillou2019meta} proposed the \md benchmark, which consists of $10$ datasets from diverse domains.
Two settings are used for reporting in general --- one where representations are learned on the train split of ``ImageNet only", and another where train sets from ``all datasets" except two are combined (the two remaining are used for testing only). After the training, few-shot evaluation is performed on the test split across all $10$ domains using many tasks by varying the ways and shots per task.

\noindent
\textbf{Instance-discriminative contrastive learning.} Among various self-supervised tasks, contrastive learning with instance discrimination as a pretext task has emerged as the leading approach for unsupervised representation learning for visual domains~\cite{dosovitskiy2015discriminative,wu2018unsupervised,chen2020simple,he2020momentum,oord2018representation}.
These methods employ various forms of data augmentations and optimize a contrastive loss that forces augmentations of the same instance to be similar in the embedding space relative to other images.
Much prior work has focused on the use of contrastive learning for pre-training, where the learned represented are evaluated on downstream tasks. 
However, sequential training may be sub-optimal due to the dynamics of training deep networks during transfer across tasks~\cite{achille2017critical}, and introducing supervision early might lead to better representations.

\noindent
\textbf{Combining supervised and self-supervised learning.}
The complementary roles of supervision and self-supervision have been explored in number of prior works. Some methods \cite{gidaris2019boosting,su2019does} use self-supervised losses (\eg, jigsaw~\cite{noroozi2016unsupervised} and rotation tasks~\cite{gidaris2018unsupervised}) as auxiliary losses during supervised training. These methods require calibrating the two losses and are not robust when combing data across domains. 
Alternate approaches combined self-supervised pre-training followed by supervised finetuning or adaptation~\cite{doersch2020crosstransformers}. We compare against these approaches.
The work most closely related to ours is \supcon~\cite{khosla2020supervised} which uses instance discrimination in a supervised setup using a modification of \simclr~\cite{chen2020simple}.
Similar to our approach they use the class labels to generate different views of the data and show superior results on ImageNet compared to standard supervised training methods.
Our work is based on \moco. While the difference between \simclr and \moco is negligible in self-supervised setting, it is significant in the supervised setting. 
In particular the queue-based architecture of \moco allows larger effective batch sizes allowing contrastive losses over diverse set of classes. Empirically we find this to be crucial for good performance.
We find that \smoco provides a 4\% improvement over \supcon on both settings on \md.
These results echo years of work in the area of metric learning that has focused on  mining triples, hard negatives, and other sampling schemes~\cite{hoffer2015deep, schroff2015facenet, hadsell2006dimensionality} to improve learning.

\noindent
\textbf{Baselines on \md.} Along with the experimental setup \cite{triantafillou2019meta} includes results with several meta-learning methods including Prototypical Networks (PN) \cite{snell2017prototypical} and MAML~\cite{finn2017model} which serve as additional baselines. 
For the ImageNet-only setup, \cite{tian2020rethinking, dhillon2019baseline} showed that a softmax classifier based supervised pre-training performs better than the meta-learning baselines. 
CrossTransformers~\cite{doersch2020crosstransformers}, the current state-of-the-art on the ImageNet-only setup, uses a self-supervised pre-training and a Transformers~\cite{vaswani2017attention} based classifier. In the all-datasets setup, current state-of-the-art methods \cite{dvornik2020selecting, liu2020universal} use a multi-task learning where a shared backbone is trained on samples from all datasets. Some domain specific parameter such as FiLM layers \cite{perez2017film} are used, which increase performance, but lead to complexity at training. At test-time, these methods use a model selection mechanism to pick the right representation to adapt to a given few-shot task.
In contrast our model trains a single network on a unified dataset created by simply aggregating images and labels from all datasets.

\section{Background}
We start by briefly describing two popular instance-discriminative contrastive learning algorithms -- \moco~\cite{he2020momentum} and \simclr~\cite{chen2020simple}, followed by describing how \supcon~\cite{khosla2020supervised} adds supervision to formulation of \simclr. 

\noindent
\textbf{\moco}~\cite{he2020momentum} is based on a contrastive loss estimated using samples in a batch $x_i, i \in \{1 \dots n\}$ and a queue ($\mathcal{Q}$) of size $K$. 
It trains two feature extractors:  a query-encoder $f_q(\cdot)$ and key-encoder $f_k(\cdot)$. Each image in the batch is transformed in two different ways $(x_i, \bar{x}_i)$ and processed through the $f_q$ and $f_k$ respectively. The contrastive loss is defined as:
\begin{align}
    \begin{split}
    \mathcal{L} &= -\log\frac{\exp(\mathcal{S}(f_q(x_i), f_k(\bar{x}_i)))}{\mathcal{D}}
    \end{split}
\end{align}
\begin{align*}
    \mathcal{D} &= \exp(\mathcal{S}(f_q(x_i), f_k(\bar{x}_i))) +\sum_{h=1}^{K} \exp(\mathcal{S}(f_q(x_i), \mathcal{Q}_h))
\end{align*}
\normalsize
where $\mathcal{Q}_h$ is the $h^{th}$ entry of the $\mathcal{Q}$ (of size $K$) and $\mathcal{S}$ is a similarity function such as the scaled cosine similarity. 
The main difference between the two encoders is that $f_q$ is updated using the gradient of the objective, while $f_k$ is updated using momentum. The encoded keys are then added to the $\mathcal{Q}$ and the oldest keys are discarded. A large value of momentum is used to ensure consistency of the keys in $\mathcal{Q}$ across batches. 

\vspace{1em}
\noindent
\textbf{\simclr}~\cite{chen2020simple} does not use any queue and estimates a contrastive loss among examples within the batch. In particular during training each batch contains $2n$ samples corresponding to two augmentations $(x_i, x_j)$ of $n$ images. The objective between a positive pair $(x_i,x_j)$  is defined as
\small
\begin{equation}
    \mathcal{L}_{ij} = -\log\frac{\exp(\mathcal{S}(f(x_i), f(x_j)))}{\sum_{k=1}^{2n}\mathbf{1}_{[k \neq i]}\exp(\mathcal{S}(f(x_i), f(x_k)))}
\end{equation}
\normalsize
where $f(\cdot)$ is a feature extractor with $f(x)=h(g(x))$ consisting of the backbone $g(\cdot)$ and a multi-layer projection head $h(\cdot)$. The overall objective consider all positive pairs in the batch.
After training $h$ is discarded and $g$ is used as the feature extractor for downstream tasks.

\vspace{1em}
\noindent
\textbf{\supcon}~\cite{khosla2020supervised} modifies the above algorithm to take into account class labels by simply considering all images from the same class along with their augmentations to be positives with respect to each other. All other images and their augmentations are considered negative. If a mini-batch contains $2n$ samples ($n$ images with one augmented view each) with $P$ unique images per class, then the loss for each $x_i$ is:

\small
\begin{equation}
    \mathcal{L} = \frac{-1}{2P-1}\sum_{r=1}^{2P-1}\log\frac{\exp(\mathcal{S}(f(x_i), f(x_r)))}{\sum_{k=1}^{2n}\mathbf{1}_{[k \neq i]}\exp(\mathcal{S}(f(x_i), f(x_k)))}
\label{eq:supcon}
\end{equation}
\normalsize
$x_r, r \in \{1 \dots 2P - 1\}$ are $2P-1$ positive samples for $x_i$ out of which one is augmented view of $x_i$ and other $2P-2$ samples are $P-1$ different images from the same class along with their augmented views. \supcon was shown to improve over standard cross-entropy based training on ImageNet as measured in terms of generalization on downstream tasks. We compare to \supcon in this work.

\section{Supervised Momentum Contrast}\label{sec: supmoco}
\smoco uses the same idea as \supcon where labels (when available) are used to define positive pairs in the \moco objective. 
The main difference is that we need to keep track of the labels in both the keys and the $\mathcal{Q}$ and consider choices of how to sample batches and update the queue.
Suppose we have sampled $B_i$ images (positives) for image $x_i$ out of which one is augmented image of $x_i$ itself and the others are random augmented images from the same class. There are $Q_i$ other samples that belong to the same class as $x_i$ in the $\mathcal{Q}$ and we denote  $B_i + Q_i$ as $P_i$.
The loss for the sample $x_i$ is:
\small
\begin{align}
    \begin{split}
         \mathcal{L} = \frac{-1}{P_i}\Bigg[&\sum_{j=1}^{B_i}\log \frac{\mathcal{S}(f_q(x_i), f_k(x_{b^i(j)}))}{\mathcal{D}} \\
         &+ \sum_{j=1}^{Q_i}\log  \frac{\mathcal{S}(f_q(x_i), \mathcal{Q}_{q^i(j)})}{\mathcal{D}}\Bigg] \\
    \end{split}
\end{align}
\begin{align*}
    \mathcal{D} &= \sum_{j=1}^{B_i} \exp(\mathcal{S}(f_q(x_i), f_k(x_{b^i(j)}))) + \sum_{h=1}^{K}\exp(\mathcal{S}(f_q(x_i), \mathcal{Q}_h))
\end{align*}
\normalsize
where $b^i(j)$ and $q^i(j)$ denote the indices of the positive samples for $x_i$ and other images belonging to the same class as $x_i$ in the queue respectively. During training, only gradient with respect to the loss of the original image $x_i$ is used to update the query-encoder $f_q$ and $f_k$ is updated using momentum instead of gradients.
We describe the details of how we sample the data and update the queue next. \footnote{PyTorch code is provided in the supplementary materials.}

\noindent
\textbf{Sample Selection:} While selecting keys (positives) for a given query image, instead of only selecting the augmented view of each image, we also sample $P-1$ additional images (data-augmented) from the training set. This allows learning representations to learn class specific invariances. We discuss the impact of the choice of $P$ in Tab.~\ref{Tab:sm-p}.

\noindent
\textbf{Queue Architecture:} Apart from storing the keys the above algorithm requires us to store the class labels as well. One choice is what to add to the queue after each batch update. We found that instead of adding all the samples in the batch to the queue, it was effective to add just one per image (the data-augmented image of each $x_i$). This increases the diversity of data-points in the queue.

\noindent
\textbf{Discussion.} By contrasting between instances within a batch leads to a tradeoff where large number of positives samples leads to a poor estimate of the denominator due to a potential lack of hard negatives. 
Decoupling the sampling strategy within the batch and queue provides a greater flexibility and larger effective batch sizes on the same GPU memory constraint. 
Empirically, we find that the performance of \supcon with a batch-size of $1024$ (maximum that fits on 8 V100 GPUs) lags behind \smoco with a  batch-size of $512$. We present the details in Sec.~\ref{sec:supvsmoco}.

\section{Experiments}
\subsection{Experimental Setup}
We describe our experimental setup below including the dataset, details regarding \smoco training and how we perform few-shot evaluation.
\subsubsection*{Dataset}
We use ~\md  \cite{triantafillou2019meta} to evaluate few-shot classification performance. ~\md  consists of $10$ datasets from different domains :  ImageNet/ILSVRC-2012 \cite{russakovsky2015imagenet}, Aircraft \cite{maji2013fine}, Omniglot \cite{lake2015human}, Textures \cite{cimpoi2014describing}, QuickDraw \cite{jongejan2016quick}, Fungi \cite{schroeder2018fgvcx}, VGG-Flower \cite{nilsback2008automated}, Birds \cite{WahCUB_200_2011}, MSCOCO \cite{lin2014microsoft} and Traffic-Sign \cite{houben2013detection}. Most of these are fine-grained datasets (\eg VGG-Flower, Aircraft, Textures, Birds, Fungi). Out of these $10$ datasets, either only the ImageNet or the first $8$ datasets can be used for training. Traffic-Sign and MSCOCO are reserved for testing only to evaluate out-of-domain generalization in case all $8$ datasets are used for training. We refer to the first setup as ``ImageNet-only'' and the second setup as ``all-datasets''. The first $8$ datasets are split into train, validation and test segments where the classes present in each segment are disjoint from each other. MSCOCO and Traffic-Sign does not have any classes belonging to the train split and therefore can not be used for training. We provide details about the datasets in the supplementary materials.

\subsubsection*{Contrastive Training Details}
We use a ResNet-18 backbone \cite{he2016deep} with $224 \times 224$ images and train using $8$ V100 GPUs (AWS P3.16XL). For \smoco, we use $3$ positive samples for every image, with one of them being the augmented view of the same image. For \supcon, we use $4$ images per class in a mini-batch which means each image gets $3$ different images and their augmentations plus its own augmentation as positives. We train for $250$ epochs when training with ImageNet-only and $300$ epochs when using all datasets. For \supcon, we train with a batch-size of $1024$ whereas for \smoco, we use a smaller batch-size of $512$. We use a linear warm-up for learning-rate during first $10$ epochs (starting from $0.1$ to peak of $0.8$ for \supcon and $0.4$ for \smoco) and train with SGD + momentum ($0.9$) with LARS \cite{you2017large} along with a weight-decay of $1e^{-4}$ and cosine-annealing learning-rate scheduling \cite{loshchilov2016sgdr}. We use $5$ data-augmentations during contrastive training: \texttt{RandomResizedCrop}, \texttt{ColorJitter}, \texttt{RandomHorizontalFlip}, \texttt{RandomGrayscale} and \texttt{GaussianBlur}. We construct the projection head with one hidden layer (with ReLU activation) of dimension $512$ and the output dimension is kept at $128$. We set the softmax-temperature $\tau$ to be $0.1$. For \smoco, we use a queue of length $16384$ and a momentum of $0.999$. When training using all the $8$ datasets, we concatenate all the training data and train using the combined dataset. While training using the combined dataset, we randomly sample images from all datasets in every mini-batch rather than ensuring that a mini-batch contains data only from a particular dataset. We provide additional results in the supplementary materials on training in this way versus keeping each mini-batch pure and show that our setup yields a much better performance. 

\subsubsection*{Few-Shot Evaluation}
 Following the protocol suggested in ~\md, we sample $600$ tasks from each of the $10$ datasets where each task contains a variable number of ways (classes) \& shots (samples) and report the average accuracy across these tasks along with the average rank across all $10$ datasets. To solve each individual task, we use a finetuning procedure similar to \cite{dhillon2019baseline} where we use a cosine classifier~\cite{gidaris2018dynamic} with the weights of the classifier initialized using the prototypes for each class (computed using the support/train set) and then finetune both the classifier and the backbone jointly. However, we do not use transductive finetuning \cite{dhillon2019baseline} to ensure a fair comparison with other methods. We use a batch-size of $64$, learning-rate of $0.001$, SGD + momentum ($0.9$) with $1e^{-4}$ weight-decay and finetune for $50$ epochs. 

\subsection{Experimental Results}
In this section, we report and analyze the performance of \smoco on both the ImageNet-only and all-datasets setup. For each segment, we provide additional details regarding existing methods and differentiate our approach against these.

\begin{table*}[t!]
\caption{Performance when trained using ImageNet-only. We use the following methods from the baselines: FS-Baseline : \textit{Transductive finetuning}; Sup. Embeddings : \textit{lr-distill}; CrossTransformers : \textit{CTX+\simclr+Aug}. \smoco outperforms all prior methods on the average rank metric and performs better on $6/10$ tasks compared to the state-of-the-art~\cite{doersch2020crosstransformers}.}
\centering{
\resizebox{1.0\textwidth}{!}{
\begin{tabular}{@{}ccccccccccccc@{}}
\toprule
\multirow{2}{*}{Algorithms} & \multirow{2}{*}{Backbone} & \multirow{2}{*}{\color[HTML]{CC338B}{Avg. Rank}} & \multicolumn{10}{c}{Test Datasets} \\ \cmidrule(l){4-13} 
 &  &  & ImageNet & Aircraft & Birds & Omniglot & Textures & MSCOCO & QuickDraw & Traffic-Sign & VGG-Flower & Fungi \\ \midrule
ProtoNets \cite{triantafillou2019meta} & ResNet-18 & \color[HTML]{CC338B}{5.75} &50.50 & 53.10 & 68.79 & 59.98 & 66.56 & 41.00 & 48.96 & 47.12 & 85.27 & 39.71 \\
Proto-MAML \cite{triantafillou2019meta} & ResNet-18 & \color[HTML]{CC338B}{5.15} & 49.53 & 55.95 & 68.66 & 63.37 & 66.49 & 43.74 & 51.52 & 48.83 & 87.15 & 39.96 \\
Sup. Embedding \cite{tian2020rethinking} & ResNet-18 & \color[HTML]{CC338B}{3.30} &61.48 & 62.32 & 79.47 & 64.31 & 79.28 & 59.28 & 60.83 & 76.33 & 91.00 & 48.53 \\
FS-Baseline \cite{dhillon2019baseline} & WRN-28-10 & \color[HTML]{CC338B}{3.25} & 60.53 & 72.40 & 82.05 & \textbf{82.07} & 80.47 & 42.86 & 57.36 & 64.37 & 92.01 & 47.72 \\
CrossTransformers \cite{doersch2020crosstransformers} & ResNet-34 & \color[HTML]{CC338B}{1.90} & \textbf{62.76} & 79.49 & 80.63 & \textbf{82.21} & 75.57 & \textbf{59.90} & \textbf{72.68} & 82.65 & \textbf{95.34} & 51.58 \\  \arrayrulecolor{black!30}\midrule

\smoco & ResNet-18 & \color[HTML]{CC338B}{\textbf{1.65}} & \textbf{62.96} & \textbf{81.48} & \textbf{84.89} & 78.42 & \textbf{88.59} & 52.18 & 68.42 & \textbf{84.69} & 93.56 & \textbf{55.39} \\ \arrayrulecolor{black}\bottomrule
\end{tabular}}}
\label{Tab:md-im}
\end{table*}

\begin{table*}[t!]
\caption{Performance when trained using all $8$ datasets of ~\md. \smoco outperforms all methods on the average rank metric and performs equal or better on $8/10$ tasks compared to the state-of-the-art~\cite{liu2020universal}. }
\centering{
\resizebox{1.0\textwidth}{!}{
\begin{tabular}{@{}ccccccccccccc@{}}
\toprule
\multirow{2}{*}{Algorithms} & \multirow{2}{*}{Backbone} & \multirow{2}{*}{\color[HTML]{CC338B}{Avg. Rank}} & \multicolumn{10}{c}{Test Datasets} \\ \cmidrule(l){4-13} 
 &  &  & ImageNet & Aircraft & Birds & Omniglot & Textures & MSCOCO & QuickDraw & Traffic-Sign & VGG-Flower & Fungi \\ \midrule
ProtoNets \cite{triantafillou2019meta} & ResNet-18 & \color[HTML]{CC338B}{6.60} & 44.50 & 71.14 & 67.01 & 79.56 & 65.18 & 39.87 & 64.88 & 46.48 & 86.85 & 40.26 \\
Proto-MAML \cite{triantafillou2019meta} & ResNet-18 & \color[HTML]{CC338B}{5.90} & 46.52 & 75.23 & 69.88 & 82.69 & 68.25 & 41.74 & 66.84 & 52.42 & 88.72 & 41.99 \\
CNAPs \cite{requeima2019fast} & ResNet-18 & \color[HTML]{CC338B}{4.90} & 52.30 & 80.50 & 72.20 & 88.40 & 58.30 & 42.60 & 72.50 & 60.20 & 86.00 & 47.40 \\
SimpleCNAPs \cite{bateni2020improved} & ResNet-18 & \color[HTML]{CC338B}{3.45} & 58.60 & 82.40 & 74.90 & 91.70 & 67.80 & 46.20 & 77.70 & 73.50 & 90.70 & 46.90 \\
SUR \cite{dvornik2020selecting} & ResNet-18 & \color[HTML]{CC338B}{3.25} & 56.30 & 85.40 & 71.40 & 93.10 & 71.50 & \textbf{52.40} & 81.30 & 70.40 & 82.80 & 63.10 \\ \arrayrulecolor{black!30}
URT \cite{liu2020universal} & N/A & \color[HTML]{CC338B}{2.35} & 55.70 & 85.80 & 76.30 & \textbf{94.40} & 71.80 & \textbf{52.20} & \textbf{82.50} & 69.40 & 88.20 & \textbf{63.50} \\ \arrayrulecolor{black!30}\midrule
\smoco & ResNet-18 & \color[HTML]{CC338B}{\textbf{1.55}} & \textbf{61.94} & \textbf{86.61} & \textbf{86.93} & 91.61 & \textbf{87.64} & 51.34 & \textbf{82.44} & \textbf{84.31} & \textbf{92.62} & \textbf{63.68} \\ \arrayrulecolor{black}\bottomrule
\end{tabular}}}

\label{Tab:md-all}
\end{table*}

\subsubsection*{Training Using ImageNet-Only}

We report the performance metrics on ImageNet-only in Tab. \ref{Tab:md-im} and compare against the following baselines:
\begin{itemize}[leftmargin=0.15in]
\setlength\itemsep{0.1em}
\item \textbf{ProtoNets/Proto-MAML :} ProtoNets (PN) trains a Prototypical Networks \cite{snell2017prototypical} on the training set using episodic sampling whereas Proto-MAML uses a first-order approximation of MAML \cite{finn2017model} where the inner (linear) classifier weights are initialized using the prototypes of every class. Both of these baselines suffer from supervision collapse. Using episodic sampling also brings an additional problem where data-points are not compared against representatives from all other classes at every training step which further affects representation quality.
\item \textbf{Supervised Embeddings/Few-Shot Baseline :} These two algorithms train an embedding using a standard supervised loss using the entire dataset without using any form of episodic sampling. Supervised Embeddings \cite{tian2020rethinking} keeps the backbone fixed at test-time and learns a Logistic Regression classifier while Few-Shot Baseline \cite{dhillon2019baseline} uses transductive finetuning. Although these methods suffer from supervision collapse, we see better performance compared to meta-learning methods because of using a $N$-way softmax classifier which ensures that each image is compared against all class representatives and creates more discriminative features.
\item \textbf{CrossTransformers :} To avoid supervision collapse, CrossTransformers \cite{doersch2020crosstransformers} proposes a self-supervised instance-discriminative pre-training phase followed by training using a Transformer based architecture which builds upon the nearest-mean classifier of PN but learns to retain the location of image features by combining features with an attention based mechanism rather than using an averaged-out feature-vector. 
\end{itemize}

Though CrossTransformers and \smoco both use the supervised datasets to train the embedding, the instance-discriminative training embedded in both the algorithms (as an initial training phase in CrossTransformers and in the single-stage training of \smoco) helps to avoid supervision collapse to a large extent and results in superior performance compared to the other baselines. 

From the experimental results reported in Tab. \ref{Tab:md-im}, we can see that \smoco outperforms all other algorithms on the average rank metric. In particular, it outperforms the current state-of-the-art CrossTransformers despite using a smaller backbone (ResNet-18 vs ResNet-34) and no additional parameters like the Transformers while also having a shorter training time due to the single-stage training mechanism.

One of the baselines that we do not compare against is using an instance-discriminative contrastive loss as an auxiliary loss similar to \cite{gidaris2019boosting}. This involves tuning a crucial hyperparameter to determine the relative importance of the standard cross-entropy loss and the self-supervised loss which requires an extensive hyperparameter search. We executed a single training run using an auxiliary instance-discriminative contrastive loss with equal weightage given to both the contrastive and supervised loss and observed that it under-performed standard supervised training \cite{tian2020rethinking, dhillon2019baseline}. 

\subsubsection*{Training Using All-Datasets}

\begin{figure}
    \centering
    \includegraphics[width=1.0\linewidth]{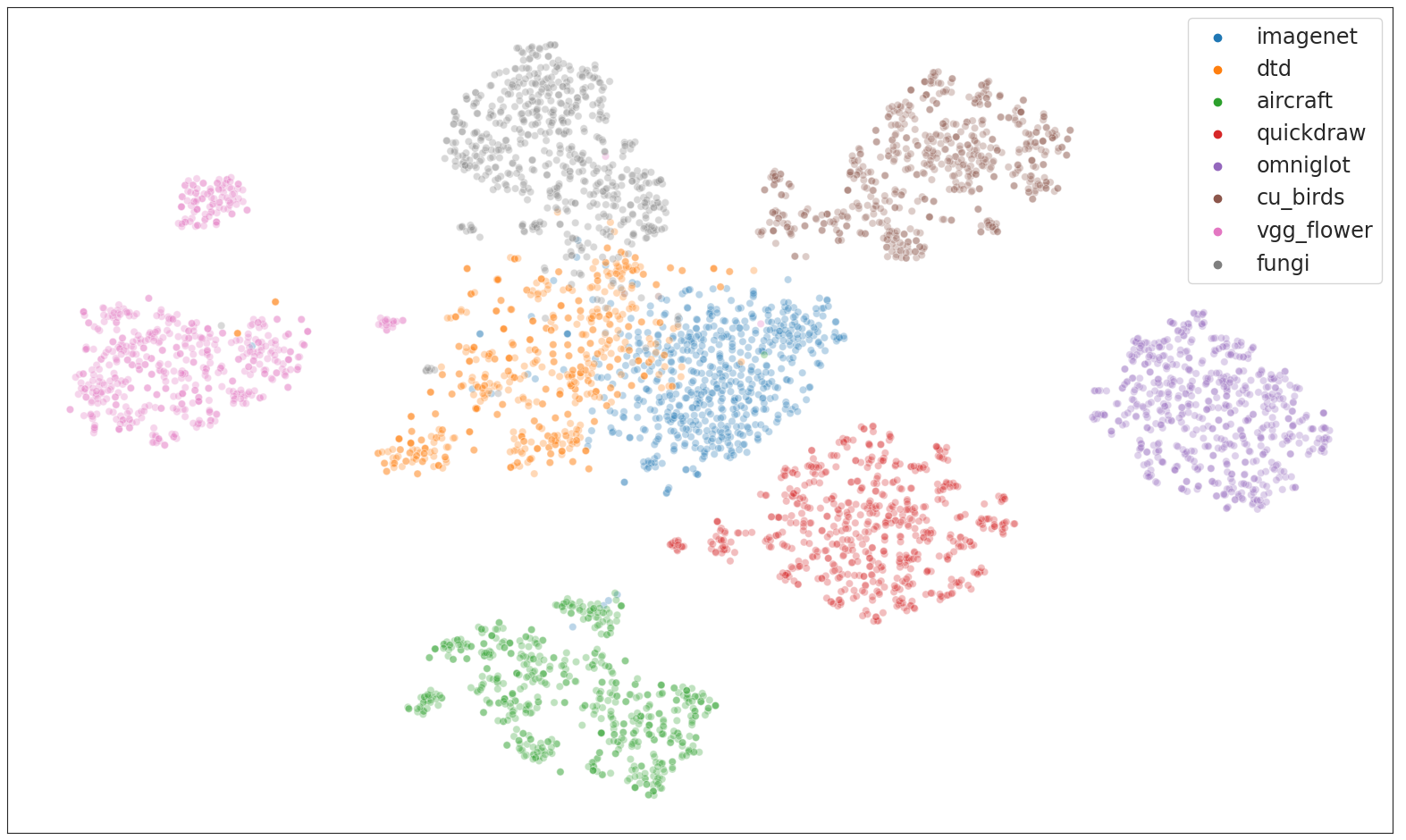}
    \caption{\small \sl t-SNE plot for visualizing the features (computed using \smoco) corresponding to the images from the validation set of all training datasets of ~\md . A single embedding is able to decipher the characteristics of individual datasets and project them onto different subspaces. This qualitatively shows that our \smoco embedding can preserve the identities of each dataset without requiring any domain specific parameters.}
    \label{fig:embedding}
\end{figure}

In this experimental setup, the algorithms can use images belonging to the train classes from all the $8$ datasets. Baseline methods can be broadly divided into three categories here -- 1) concatenate all data (and labels) and train using it 2) train a common backbone and one additional set of parameter to adapt based on the domain 3) train a common backbone and a set of additional parameters per domain and use a model selection mechanism at test-time. Previously discussed baselines (PN/Proto-MAML) use the first approach. In \smoco as well, we take the first approach of training a single model (no domain specific parameters) by concatenating data from all the classes across $8$ domains.

\begin{itemize}[leftmargin=0.15in]
\setlength\itemsep{0.1em}
\item \textbf{CNAPs/SimpleCNAPs :} CNAPS/SimpleCNAPS are few-shot classifiers based on conditional neural processes (CNP). Both use a shared feature extractor with one set of FiLM~\cite{perez2017film} layers that is adapted using meta-learning. CNAPS uses a linear classifier while SimpleCNAPS uses a Mahalanobis distance classifier to solve each task.

\item \textbf{SUR/URT :} SUR~\cite{dvornik2020selecting} and URT~\cite{liu2020universal} use the idea of universal representations~\cite{bilen2017universal} where a shared backbone along with domain specific parameters (implemented via FiLM layers) is used for training. The idea is to share information across domains while also retaining individual properties of each domain via a few domain specific parameters. At test-time, SUR uses a probabilistic model to find out how individual domain representations should be combined given a target task. On the other hand, URT meta-trains a Transformer layer (after the universal backbone is trained) for learning-to-learn such a combination. Both of these methods can be considered a form of ``soft'' model selection as opposed to a ``hard'' selection where features corresponding to one particular domain is picked.
\end{itemize}

From the experimental results reported in Tab. \ref{Tab:md-all}, we can see that \smoco outperforms all other algorithms on the average rank metric. It may seem surprising that \smoco can outperform other methods, especially the ones which use domain specific parameters. However, if we see the \smoco embedding space (Fig \ref{fig:embedding}), we can see that it preserves the individuality of each domain without requiring any domain specific parameters. Using domain specific parameters comes with an additional downside of having to use model selection at test time. Given the limited amount of labeled data available during few-shot testing, the selection process may get biased and assign more importance to the parameters corresponding to an unrelated domain. Having a single embedding alleviates that problem as the embedding itself possesses all the information across domains and can be adapted to the target task as required. 

\subsection{Additional Experiments}

\begin{figure*}
    \includegraphics[width=1.0\linewidth]{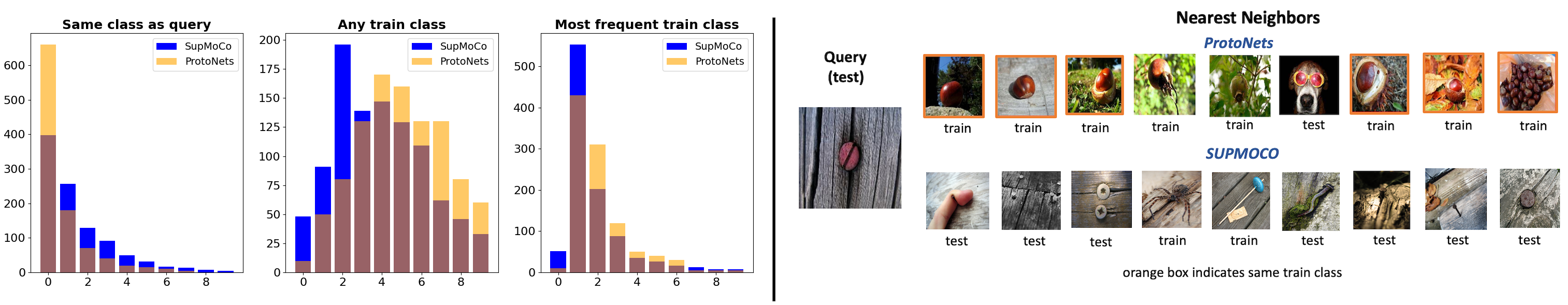}
    \caption{\small In the left side, we show the quantitative results of the supervision collapse experiment (Sec.~\ref{sec:sup-collapse}). On the leftmost plot, X-axis shows the number of retrievals from the same (test) class ($0$ means none from same test class). In the middle plot, it indicates number of retrievals from train classes ($0$ indicates all from test). In the rightmost plot, X-axis denotes the maximum frequency of the same train class \eg a value of $3$ means at max 3 members were from the same train class ($0$ means all from test). The Y-axis value denotes the number of queries in each bin. In the leftmost plot, \smoco can be seen to shift the mass to the right which means it can find more samples with the same class as the test image in the retrieval space. On middle and rightmost plot, \smoco shifts the mass to the left which means it matches the test image less with images from any train class and a particular train class respectively and generates more differentiating features for unseen images. On the right side, we show the nearest neighbors of a particular test image retrieved using the two algorithms. We see the representation collapse with ProtoNets where the features for the query image ends up being similar to the ``buckeye'' class from training because the network associates the red circle of the query image to the train class and ignores other contextual information. In comparison, \smoco understands the visual semantics of the image better and finds images predominantly from the test set which are very similar to the query image (``screws'' and ``nails'' in the wild).}
    \label{fig:collapse}
\end{figure*}

\subsubsection*{Analyzing Supervision Collapse}\label{sec:sup-collapse}
Standard supervised training methods suffer from supervision collapse where they discard information which is not relevant for the immediate training task. In this experiment,  we use the experimental setup provided by \cite{doersch2020crosstransformers} to analyze ``supervision collapse'' both qualitatively and quantitatively between a supervised meta-learning algorithm (ProtoNets~\cite{snell2017prototypical}) and \smoco. The experimental setup is based on performing nearest-neighbor (NN) retrievals in the joint embedding space (train + test) of the ImageNet dataset. The retrieval set is constructed by sampling $130$ random images from each of the $712$ train class and $130$ test classes. The task is to find the top-$9$ nearest neighbors for $1000$ randomly sampled images in this joint embedding space and evaluate : 
\begin{itemize}[leftmargin=0.15in]
\setlength\itemsep{0.1em}
\item Number of NNs that come from the same (test) class as the test/query image. 
\item Number of NNs that come from the train set.
\item Among NNs that come from the same train class, number of most frequently-retrieved such class. 
\end{itemize}
The first metric analyzes how differentiated the representation of each test class is while the second metric measures how much the representation of an individual test image collapses to some image from the train set. The third metric evaluates collapsing on one train class -- if a majority of the retrieval comes from one particular train class, it indicates that the representation of that image has predominantly coincided with that train class representation. 

The empirical analysis is reported in the left side of Figure \ref{fig:collapse}. We can observe that \smoco has at least one neighbor from the same test class in $60\%$ of the cases while it is $34.1\%$ for PN (as reported in ~\cite{doersch2020crosstransformers}, higher the better). When it comes to evaluating collapse on the same (train) class, \smoco has $39\%$ cases where two or more neighbors are from the train class while it is $55.3\%$ for PN (lower the better). This indicates that \smoco prevents collapse better than PN and generates more unique representations for unseen images. Qualitative analysis (Fig.~\ref{fig:collapse} right) shows similar findings as the quantitative one.

\begin{table*}[t!]
\caption{Performance comparison among \smoco trained with ImageNet-only (\smoco-IM), with ImageNet and $10\%$ of the labeled data (rest provided as unlabeled) from other domains (\smoco-SSL) and with all datasets (\smoco). While only using $10\%$ of the data, \smoco only has an average of $2\%$ performance gap compared to the fully supervised model. When comparing with the model trained on ImageNet alone, \smoco-SSL can achieve $4\%$ gain on domains distant from ImageNet (\eg Fungi, Omniglot which are indicated in blue).}
\centering{
\small
\resizebox{1.0\textwidth}{!}{
\begin{tabular}{lllllllllll}
\hline
 & \multicolumn{10}{c}{Test Datasets} \\ \cline{2-11} 
\multirow{-2}{*}{Data} & \multicolumn{1}{l}{ImageNet} & Aircraft & Birds & Omniglot & Textures & MSCOCO & QuickDraw & Traffic-Sign & VGG-Flower & Fungi \\ \hline
\smoco-IM & 62.96 & \cellcolor[HTML]{34CDF9}{\color[HTML]{000000}81.48} & 84.89 & \cellcolor[HTML]{34CDF9}{\color[HTML]{000000}78.42} & 88.59 & 52.18 & \cellcolor[HTML]{34CDF9}{\color[HTML]{000000}68.42} & 84.69 & 93.56 & \cellcolor[HTML]{34CDF9}{\color[HTML]{000000}55.39} \\ 
\smoco-SSL & 61.92 & \cellcolor[HTML]{34CDF9}{\color[HTML]{000000}83.41} & 85.09 & \cellcolor[HTML]{34CDF9}{\color[HTML]{000000}86.17} & 86.97 & 51.21 & \cellcolor[HTML]{34CDF9}{\color[HTML]{000000}79.93} & 84.35 & 92.43 & \cellcolor[HTML]{34CDF9}{\color[HTML]{000000}59.67} \\ 
\smoco & 61.94 & \cellcolor[HTML]{34CDF9}{\color[HTML]{000000}86.61} & 86.93 & \cellcolor[HTML]{34CDF9}{\color[HTML]{000000}91.61} & 87.64 & 51.34 & \cellcolor[HTML]{34CDF9}{\color[HTML]{000000}82.44} & 84.31 & 92.62 & \cellcolor[HTML]{34CDF9}{\color[HTML]{000000}63.68}  \\ \hline
\end{tabular}}}
\label{Tab:ssl}
\end{table*}

\begin{table*}[t!]
\caption{Performance comparison between \supcon and \smoco on ImageNet-only. \smoco outperforms \supcon by $3.6\%$ on average across all the tasks.}
\centering{
\small
\resizebox{1.0\textwidth}{!}{
\begin{tabular}{lllllllllll}
\hline
 & \multicolumn{10}{c}{Test Datasets} \\ \cline{2-11} 
\multirow{-2}{*}{Algorithms} & \multicolumn{1}{l}{ImageNet} & Aircraft & Birds & Omniglot & Textures & MSCOCO & QuickDraw & Traffic-Sign & VGG-Flower & Fungi \\ \hline
\supcon & 59.30 & 78.39 & 81.86 & 74.60 & 84.88 & 48.36 & 64.31 & 81.23 & 90.16 & 51.41 \\ 
\smoco & \textbf{62.96} & \textbf{81.48} & \textbf{84.89} & \textbf{78.42} & \textbf{88.59} & \textbf{52.18} & \textbf{68.42} & \textbf{84.69} & \textbf{93.56} & \textbf{55.39} \\ \hline
\end{tabular}}}
\label{Tab:cs-im}
\end{table*}

\begin{table*}[t!]
\caption{Performance comparison between \supcon and \smoco on all-datasets. \smoco outperforms \supcon by $4.1\%$ on average across all the tasks.}
\centering{
\small
\resizebox{1.0\textwidth}{!}{
\begin{tabular}{lllllllllll}
\hline
 & \multicolumn{10}{c}{Test Datasets} \\ \cline{2-11} 
\multirow{-2}{*}{Algorithms} & \multicolumn{1}{l}{ImageNet} & Aircraft & Birds & Omniglot & Textures & MSCOCO & QuickDraw & Traffic-Sign & VGG-Flower & Fungi \\ \hline
\supcon & 56.50 & 83.20 & 83.70 & 86.80 & 82.02 & 47.89 & 78.09 & 81.23 & 89.09 & 59.57 \\ 
\smoco & \textbf{61.94} & \textbf{86.61} & \textbf{86.93} & \textbf{91.61} & \textbf{87.64} & \textbf{51.34} & \textbf{82.44} & \textbf{84.31} & \textbf{92.62} & \textbf{63.68} \\ \hline
\end{tabular}}}
\label{Tab:cs-all}
\end{table*}

\begin{table*}[t!]
\caption{Comparison between $1$ positive per image (augmented view of the same image) and $3$ positives per image (1 augmented view + 2 random augmented images from the same class) when training \smoco using ImageNet-only. Using additional positives beyond the augmented view of itself helps to provide additional performance gain.}
\centering{
\small
\resizebox{1.0\textwidth}{!}{
\begin{tabular}{lllllllllll}
\hline
 & \multicolumn{10}{c}{Test Datasets} \\ \cline{2-11} 
\multirow{-2}{*}{Positives ($P$)} & \multicolumn{1}{l}{ImageNet} & Aircraft & Birds & Omniglot & Textures & MSCOCO & QuickDraw & Traffic-Sign & VGG-Flower & Fungi \\ \hline
$P=1$ & 60.77 & 81.34 & 80.03 & 77.16 & 86.74 & 47.19 & 64.43 & 82.35 & 91.95 & 53.56 \\
$P=3$ & \textbf{62.96} & \textbf{81.48} & \textbf{84.89} & \textbf{78.42} & \textbf{88.59} & \textbf{52.18} & \textbf{68.42} & \textbf{84.69} & \textbf{93.56} & \textbf{55.39} \\ \hline
\end{tabular}}}
\label{Tab:sm-p}
\end{table*}

\subsubsection*{Comparing \smoco and \supcon}\label{sec:supvsmoco}
Performance of \supcon improves when it has more number of positive samples per class within every mini-batch \cite{khosla2020supervised}. However, increasing number of samples per class reduces number of unique classes within a batch. This is less of a problem for self-supervised \simclr because each image is considered its own class but \supcon uses the true class labels of each image and therefore, number of distinct classes reduce by a factor of $P$ when there are $P$ samples per class. This problem does not exist for \smoco because it uses a separate queue to store features corresponding to the negative samples. This ensures that representations from all classes are available to compare against at every step even with a small batch-size. For example, in \supcon, with a batch-size of $1024$ and $4$ samples per class, we can only have $1024/4=256$ unique classes to compare against. Whereas in \smoco, irrespective of the batch-size, a queue of moderate size (\eg $8192$) can store enough samples from all classes in the larger all-datasets training setup. The queue in \smoco only has to store low dimensional feature vectors ($128$) rather than the image itself (and its features) and therefore has negligible GPU memory overhead if queue size increases. The ability to easily compare against representations from \underline{all} classes at every training step helps \smoco to produce more discriminative features. 
From the results in Tab. \ref{Tab:cs-im} \& \ref{Tab:cs-all}, we can observe that \smoco outperforms \supcon by $3.7\%$ on ImageNet-only and $4.2\%$ on all-datasets (on average) while achieving a maximum gain of $5.6\%$.

\subsubsection*{Using Partially Labeled Data}
In this experiment, we measure the performance of \smoco in a partially labeled (semi-supervised setup) to evaluate 1) its flexibility to work with both fully and partially labeled data and 2) its ability to use only a limited number of labeled samples to learn class semantics while predominantly using unlabeled data to learn individual characteristics of data from different domains. In this setup, we provide all labeled images from ImageNet and only $10\%$ labeled images from the remaining $7$ datasets while the remaining images are provided without labels. Depending on the dataset, this can provide as few as $1$ sample for certain classes. The goal here is to see how much performance gap there is between \smoco with partially labeled against all labeled data and also how much its performance improves compared to training using ImageNet-only. From the results in Tab~\ref{Tab:ssl}, we can see that the performance using $10\%$ of labeled data from the $7$ domains has only $2\%$ performance gap on average compared to the fully-supervised model. When compared to ImageNet-only, we see $4\%$  performance gain (on average) in domains which are further from ImageNet (\eg Omniglot, Aircraft, QuickDraw and Fungi) while on other domains performance stays largely same.

\subsubsection*{Choosing Additional Positives in \smoco}\label{sec:sm-p}
In \smoco, during every image, we sample $P$ positive samples for each class including one augmented view of the image. However, each sample also has some positive samples from the queue itself. In this experiment, we empirically evaluate adding extra positive samples from the same class. From Tab. \ref{Tab:sm-p}, we can see that using these additional samples help to provide some performance benefit and we hypothesize that it happens because these additional positives are encoded using the latest version of the key encoder and provides a more accurate estimate of the features to compute similarity against unlike the ones coming from the queue carrying slightly outdated features.

\section{Conclusion}
In this work, we show that combining self-supervised instance-discriminative contrastive training with supervision can perform favorably on cross-domain few-shot recognition tasks. Our proposed algorithm \smoco can outperform prior methods on \md and also performs better than a similar method called \supcon. \smoco also offers additional flexibility to use partially labeled datasets because of how it incorporates supervision and self-supervision into the algorithm. Our approach provides a new direction for improving few-shot classification by leveraging instance-discriminative contrastive learning in both supervised and semi-supervised meta-learning setup and we hope to see future works exploring this further.

{\small
\balance
\bibliographystyle{ieee_fullname}
\bibliography{egbib}
}

\appendix
\clearpage
\onecolumn

\section*{\hfil Supplementary Materials \hfil}

\section{\smoco vs \moco}
While our few-shot evaluation setup provides a large labeled dataset for pre-training, we still want to investigate the usefulness of labels by comparing against a model trained using only the images (without labels). We train a self-supervised \moco~\cite{he2020momentum} model on both the ImageNet-only setup and all-datasets setup in an unsupervised fashion to compare performance against a \smoco model. From Tab.~\ref{Tab:moco-im} and ~\ref{Tab:moco-all}, we can observe that using labels indeed helps to boost performance on few-shot tasks across all domains, with gains as large as $14.5\%$. We argue this happens because the self-supervised representation predominantly learns mid and low-level features~\cite{zhao2020makes} and does not capture enough high-level semantics. Such an embedding would be adequate for transferring to a downstream task which has a moderate number of labeled samples because it can learn the (missing) high-level representations using the available supervision. However, in a few-shot setup, label information is limited and there are not enough opportunities to learn high-level features that is required to distinguish a class from another in any classification setup. By performing this experiment, we show that using supervision with the instance discriminative learning paradigm is more helpful in a few-shot classification setup and can outperform a self-supervised model significantly. 

\section{Maintaining Data Purity within a Batch}
When training a single \smoco model on the combined dataset (\eg training on all $8$ datasets of \md), there are two ways to construct a mini-batch - keep each batch \textit{pure} by making it contain images only from a particular dataset or make it \textit{impure} by not making any dataset specific delineation and make every batch contain random samples from all the datasets. In our experimental section, we mentioned that in such a multi-domain training scenario, we use the impure batch approach because it performs better. In Tab.~\ref{Tab:moco-batch}, we compare the performance between using pure vs impure batch in details and show that impure batch outperforms pure batch across tasks from all domains. We hypothesize this to happen because in impure batch setup, \texttt{batch\_norm} parameters face lesser interference and sudden change compared to pure batch where every batch would present a drastically different set of images and cause large updates to the parameters, thereby making the training process sub-optimal. 

\section{Confidence Interval Results}
In Tab.~\ref{Tab:moco-ci}, we provide the confidence intervals when \smoco models were evaluated using $600$ few-shot randomly sampled few-shot tasks from each domain in both ImageNet-only and all-datasets setup. Because there is inherent randomness in task sampling, this helps to make a fair comparison across methods while calculating the average rank metric.

\section{Dataset Details}
In this section, we provide a detailed description of \md \cite{triantafillou2019meta}. It consists of $10$ datasets from different domains which we will describe next. Each dataset is divided into a set of disjoint training, validation and test classes and we are only allowed to train using images corresponding to the training splits from $8$ of these datasets. The other $2$ datasets are reserved for testing only.
\begin{itemize}[leftmargin=0.15in]
\setlength\itemsep{0.1em}
    \item ImageNet/ILSVRC-2012 \cite{russakovsky2015imagenet} : ImageNet is a dataset of $1000$ classes containing natural images which are split into $712-158-130$ for training-validation-test.
    \item Aircraft \cite{maji2013fine} : Aircraft is a fine-grained dataset of aircraft images which are split into $70-15-15$ for training-validation-test. All images are cropped using the bounding box information associated with each image. 
    \item Omniglot \cite{lake2015human} : Omniglot is a dataset of images of handwritten characters divided into $1623$ classes from $50$ different alphabet classes. $1623$ classes are split into $883-81-659$ for training-validation-test.
    \item Textures \cite{cimpoi2014describing} : It is a collection of texture images in the wild and the dataset is split into $33-7-7$ classes for training-validation-test.
    \item QuickDraw \cite{jongejan2016quick} : QuickDraw is a dataset of $50$ million doodle drawings across $345$ categories which is divided into $241-52-52$ categories for training-validation-test. For this dataset, we only use $2000$ samples per class to speed up training time. 
    \item Fungi \cite{schroeder2018fgvcx} : It is a fine-grained dataset containing over $100000$ fungi images and classes are split into $994-200-200$ for training-validation-test.
    \item VGG-Flower \cite{nilsback2008automated} : It is a dataset of natural images of flowers and split into $71-15-16$ for training-validation-test.
    \item Birds \cite{WahCUB_200_2011} : A dataset for fine-grained classification of $200$ bird species and the classes are split into $140-30-30$ for training-validation-test.
    \item MSCOCO \cite{lin2014microsoft} : MSCOCO is a popular object detection dataset containing 1.5 million objects across $80$ classes. For this task, individual images are extracted by cropping using the bounding box associated with each object. This dataset does not allow any images to be used for training and $80$ classes are split into $40-40$ for validation and testing.
    \item Traffic-Sign \cite{houben2013detection} : It is a dataset of $50000$ images of traffic signs across $43$ classes and the entire dataset is reserved for testing only. 
\end{itemize}

\begin{table*}[t!]
\caption{Performance comparison between \moco and \smoco when trained using ImageNet-only. \smoco clearly outperforms \moco on tasks from all domains, with an average difference of $7.5\%$.}
\centering{
\small
\resizebox{1.0\textwidth}{!}{
\begin{tabular}{lllllllllll}
\hline
 & \multicolumn{10}{c}{Test Datasets} \\ \cline{2-11} 
\multirow{-2}{*}{Batch Type} & \multicolumn{1}{l}{ImageNet} & Aircraft & Birds & Omniglot & Textures & MSCOCO & QuickDraw & Traffic-Sign & VGG-Flower & Fungi \\ \hline
\moco & 55.35 & 78.09 & 70.31 & 74.51 & 82.51 & 44.20 & 58.56 & 80.22 & 90.02 & 50.23 \\
\smoco & \textbf{62.96} & \textbf{81.48} & \textbf{84.89} & \textbf{78.42} & \textbf{88.59} & \textbf{52.18} & \textbf{68.42} & \textbf{84.69} & \textbf{93.56} & \textbf{55.39}  \\ \hline
\end{tabular}}}
\label{Tab:moco-im}
\end{table*}

\begin{table*}[t!]
\caption{Performance comparison between \moco and \smoco when trained using all-datasets. \smoco does better than \moco on all domains here as well, with an average performance gap of $6.5\%$.}
\centering{
\small
\resizebox{1.0\textwidth}{!}{
\begin{tabular}{lllllllllll}
\hline
 & \multicolumn{10}{c}{Test Datasets} \\ \cline{2-11} 
\multirow{-2}{*}{Batch Type} & \multicolumn{1}{l}{ImageNet} & Aircraft & Birds & Omniglot & Textures & MSCOCO & QuickDraw & Traffic-Sign & VGG-Flower & Fungi \\ \hline
\moco & 53.96 & 79.48 & 69.61 & 83.71 & 83.93 & 43.03 & 70.02 & 80.20 & 91.29 & 53.89  \\
\smoco & \textbf{61.94} & \textbf{86.61} & \textbf{86.93} & \textbf{91.61} & \textbf{87.64} & \textbf{51.34} & \textbf{82.44} & \textbf{84.31} & \textbf{92.62} & \textbf{63.68} \\ \hline
\end{tabular}}}
\label{Tab:moco-all}
\end{table*}

\begin{table*}[t!]
\caption{Performance comparison when a batch contains sample from all datasets (Impure Batch) vs only from a particular dataset (Pure Batch) during \smoco training.}
\centering{
\small
\resizebox{1.0\textwidth}{!}{
\begin{tabular}{lllllllllll}
\hline
 & \multicolumn{10}{c}{Test Datasets} \\ \cline{2-11} 
\multirow{-2}{*}{Batch Type} & \multicolumn{1}{l}{ImageNet} & Aircraft & Birds & Omniglot & Textures & MSCOCO & QuickDraw & Traffic-Sign & VGG-Flower & Fungi \\ \hline
Pure Batch & 50.60 & 76.39 & 69.81 & 78.24 & 77.01 & 43.36 & 75.78 & 85.73 & 85.98 & 48.41 \\ 
Impure Batch & \textbf{61.94} & \textbf{86.61} & \textbf{86.93} & \textbf{91.61} & \textbf{87.64} & \textbf{51.34} & \textbf{82.44} & \textbf{84.31} & \textbf{92.62} & \textbf{63.68}  \\ \hline
\end{tabular}}}
\label{Tab:moco-batch}
\end{table*}

\begin{table*}[t!]
\caption{Confidence interval when the \smoco models trained using ImageNet-only and all-datasets respectively were evaluated on $600$ few-shot tasks from each domain.}
\centering{
\small
\resizebox{1.0\textwidth}{!}{
\begin{tabular}{lllllllllll}
\hline
 & \multicolumn{10}{c}{Test Datasets} \\ \cline{2-11} 
\multirow{-2}{*}{Dataset} & \multicolumn{1}{l}{ImageNet} & Aircraft & Birds & Omniglot & Textures & MSCOCO & QuickDraw & Traffic-Sign & VGG-Flower & Fungi \\ \hline
ImageNet-only & $62.96\pm1.09$ & $81.48\pm1.42$ & $84.89\pm0.84$ & $78.42\pm1.40$ & $88.59\pm0.82$ & $52.18\pm1.03$ & $68.42\pm1.12$ & $84.69\pm1.35$ & $93.56\pm0.62$ & $55.39\pm1.32$  \\
All-Datasets & $61.94\pm1.04$ & $86.61\pm0.83$ & $86.93\pm0.74$ & $91.61\pm0.65$ & $87.64\pm0.93$ & $51.34\pm1.02$ & $82.44\pm0.58$ & $84.31\pm0.98$ & $92.62\pm0.76$ & $63.68\pm1.12$  \\ \hline
\end{tabular}}}
\label{Tab:moco-ci}
\end{table*}

\section{PyTorch Code}
In Alg.~\ref{Alg: Sup-MoCo}, we provide a PyTorch implementation sketch of the \smoco algorithm that was used for the fully-supervised setup. For the semi-supervised setup, the code was similar with the major difference being --- we only find positive entries from the queue corresponding to those images for which we have label information available. For others, we treat all queue elements as negative. 

\centering
\begin{minipage}{.7\linewidth}
\begin{algorithm}[H]
\caption{\smoco (PyTorch skeleton code)}
\label{Alg: Sup-MoCo}
\SetAlgoLined
\begin{lstinputlisting}[language=Python]{sup_moco.py}
\end{lstinputlisting}
\end{algorithm}
\end{minipage}

\end{document}